%% file: main.tex
\documentclass[10pt,twocolumn,letterpaper]{article}

\usepackage{iccv}              


\definecolor{iccvblue}{rgb}{0.21,0.49,0.74}
\usepackage[pagebackref,breaklinks,colorlinks,allcolors=iccvblue]{hyperref}
\usepackage{bm}
\usepackage{times}
\usepackage{epsfig}
\usepackage{graphicx}
\usepackage{amsmath}
\usepackage{amssymb}
\usepackage{microtype}
\usepackage{xcolor}
\usepackage{booktabs}
\usepackage{makecell}
\usepackage{caption}
\usepackage{tabularx}
\usepackage{booktabs}
\usepackage{multirow}

\def\paperID{9549} 
\def\confName{ICCV}
\def\confYear{2025}

\title{Enabling Versatile Controls for Video Diffusion Models}

\author{%
  Xu Zhang\textsuperscript{1},
  Hao Zhou\textsuperscript{1},
  Haoming Qin\textsuperscript{1,2},
  Xiaobin Lu\textsuperscript{1,3},
  \\
  Jiaxing Yan\textsuperscript{1},
  Guanzhong Wang\textsuperscript{1},
  Zeyu Chen\textsuperscript{1},
  Yi Liu\textsuperscript{1}
  \\
  \\
  \textsuperscript{1} PaddlePaddle Team, Baidu Inc.,
  \textsuperscript{2} Xiamen University,
  \textsuperscript{3} Sun Yat-sen University,
  \\
  \\
  \textbf{\url{https://pp-vctrl.github.io}}
  \\
}  

\def\para#1{\vspace{0.25em}\noindent\textbf{#1}}

\begin{document}

\twocolumn[{
\maketitle
\begin{center}
    \captionsetup{type=figure}
    \includegraphics[width=0.9\linewidth]{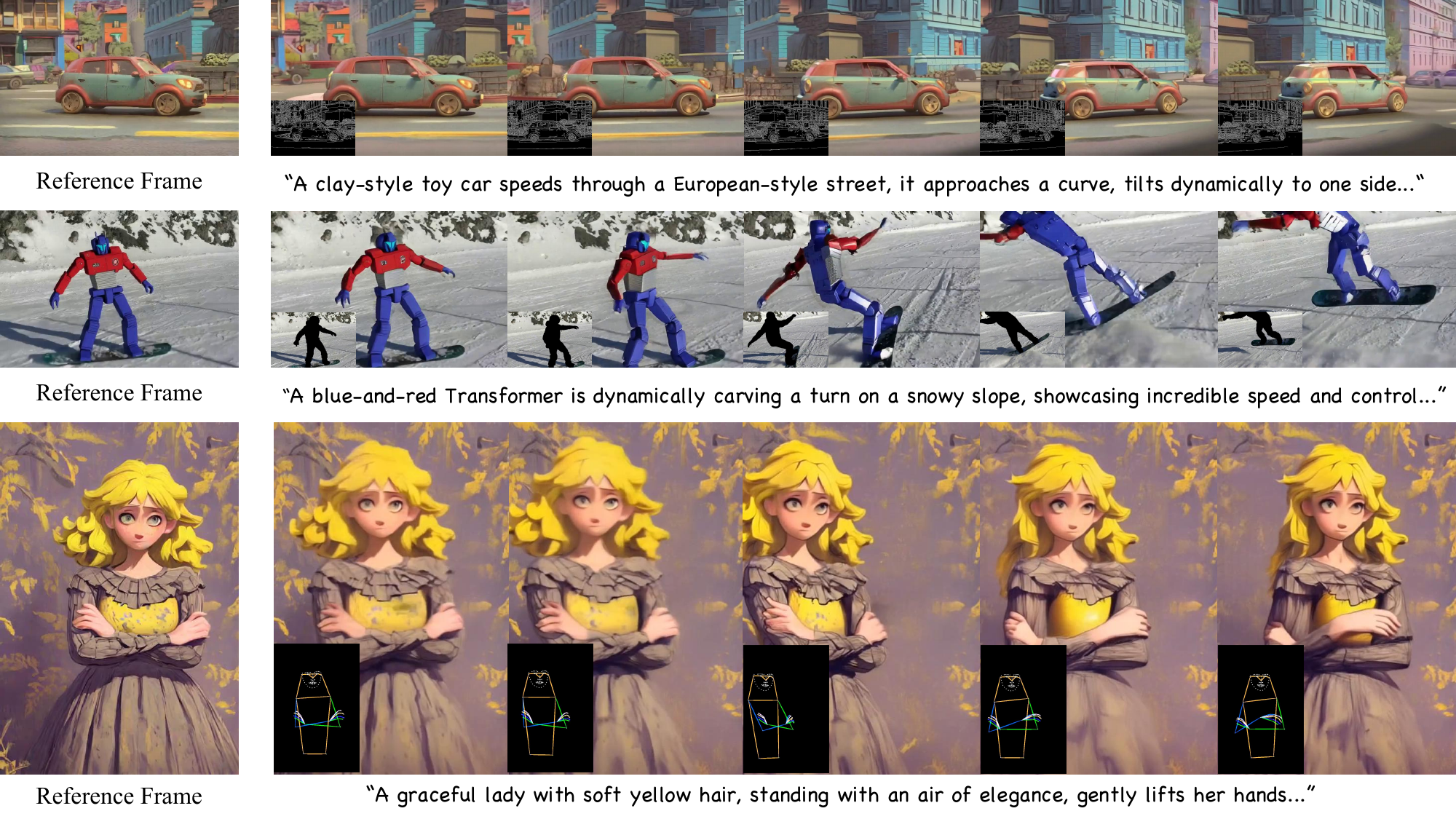}
    \vspace{-1em}
    \captionof{figure}{\textbf{Examples generated by VCtrl (also termed PP-VCtrl) using reference frames and text prompts.} VCtrl enables users to guide large pretrained video diffusion models using diverse controls, including Canny edges (top), segmentation masks (middle), and human keypoints (bottom), generating high-quality videos that accurately adhere to the provided control signals.}
    \label{fig:summary}
\end{center}
}]

\input{sec/0_abstract}    
\input{sec/1_intro}
\input{sec/2_related_work}

\input{sec/3_method}

\input{sec/4_experiments}
\input{sec/5_conclusion}
{
    \small
    \bibliographystyle{ieeenat_fullname}
    \bibliography{main}
}

\end{document}

%% file: sec/0_abstract.tex
\begin{abstract}
Despite substantial progress in text-to-video generation, achieving precise and flexible control over fine-grained spatiotemporal attributes remains a significant unresolved challenge in video generation research. To address these limitations, we introduce VCtrl (also termed PP-VCtrl), a novel framework designed to enable fine-grained control over pre-trained video diffusion models in a unified manner. VCtrl integrates diverse user-specified control signals—such as Canny edges, segmentation masks, and human keypoints—into pretrained video diffusion models via a generalizable conditional module capable of uniformly encoding multiple types of auxiliary signals without modifying the underlying generator. Additionally, we design a unified control signal encoding pipeline and a sparse residual connection mechanism to efficiently incorporate control representations. Comprehensive experiments and human evaluations demonstrate that VCtrl effectively enhances controllability and generation quality. 
The source code and pre-trained models are publicly available and implemented using the PaddlePaddle framework at \url{https://github.com/PaddlePaddle/PaddleMIX/tree/develop/ppdiffusers/examples/ppvctrl}.
\end{abstract}

%% file: sec/1_intro.tex
\section{Introduction}
\label{sec:intro}
Significant advancements in text-to-video diffusion models~\cite{he2022latent,karras2023dreampose, ruan2023mm,  openai2024sora, yang2024cogvideox, kong2024hunyuanvideo} have revolutionized video creation and editing by enabling automatic synthesis from natural language descriptions. Despite these advances, existing text-driven methods often struggle to achieve precise control over fine-grained spatiotemporal elements, such as motion trajectories, temporal coherence, and scene transitions. This limitation typically necessitates iterative and inefficient prompt engineering to achieve desired results. To address these challenges, researchers have explored the use of supplementary conditioning signals, including structural cues~\cite{chen2023controlavideo, zi2024cococo, text2video-zero}, motion data~\cite{wu2024draganything, ling2024motionclone, wang2024motionctrl, jeong2024vmc}, and geometric information~\cite{fu20243dtrajmaster, qiu2024freetraj, zhang2024tora}. However, current research predominantly adopts task-specific approaches, such as human image animation~\cite{ma2024follow, hu2024animate}, text-guided inpainting~\cite{zi2024cococo}, and motion-guided generation~\cite{wang2024motionctrl, li2024image, yin2023dragnuwa}, leading to fragmented methodologies and limited cross-task flexibility. This fragmentation highlights the need for more unified frameworks that can generalize across diverse video synthesis tasks while maintaining fine-grained control over spatiotemporal dynamics.

Several unified frameworks have emerged to address these limitations, yet critical challenges remain unresolved. First, existing methods like Text2Video-Zero~\cite{khachatryan2023text2video}, Control-A-Video~\cite{chen2023control}, and Videocomposer~\cite{wang2023videocomposer} primarily adapt image generation models instead of architectures explicitly designed for video generation, resulting in compromised temporal coherence and visual quality. Second, unlike the image domain where ControlNet~\cite{zhang2023adding} provides a unified and extensible control framework, current video generation approaches~\cite{chen2023control, wang2023videocomposer} remain tightly coupled to particular base models, restricting their scalability and broader applicability. Third, despite abundant raw video data, the lack of effective preprocessing and filtering strategies has resulted in a scarcity of high-quality datasets for controllable video generation. 
Collectively, these issues contribute to a fragmented methodological landscape, impeding progress toward a generalizable and unified framework for controllable video generation.

In this work, we introduce VCtrl (also termed PP-VCtrl), a novel architecture designed to 
enable fine-grained control over pre-trained video diffusion models in a unified manner. 
Our approach introduces an auxiliary conditioning module while maintaining the 
original generator's architecture intact. 
By leveraging the rich representations learned during pretraining, VCtrl achieves flexible control across diverse conditioning signals with minimal additional computational overhead.
Our approach features a unified control signal encoding process that transforms diverse conditioning inputs into a unified representation, while incorporating task-aware masks to enhance adaptability across different applications.
The integration with the base network is accomplished through sparse residual connection mechanism, facilitating controlled feature propagation while maintaining computational efficiency.
Additionally, we develop an efficient data filtering pipeline leveraging advanced preprocessing techniques, recaptioning methods, and task-aware annotation strategies to substantially enhance semantic alignment and overall video generation quality.

In summary, VCtrl addresses the fragmented, domain-specific landscape in controllable video generation by providing a unified, generalizable framework. Our method:
(1) enables unified control over video diffusion models through a generalizable architecture that handles multiple control types through a
conditional module without modifying the base generator; 
(2) features a unified control signal encoding pipeline combined with sparse residual connections, complemented by an efficient data filtering pipeline, enabling precise spatiotemporal control with high computational efficiency; 
and (3) demonstrates comparable or superior performance to specialized task-specific methods across various controllable video generation tasks, as validated through comprehensive experiments and user studies.

%% file: sec/2_related_work.tex
\begin{figure*}
    \vspace{-10pt}
    \includegraphics[width=\linewidth]{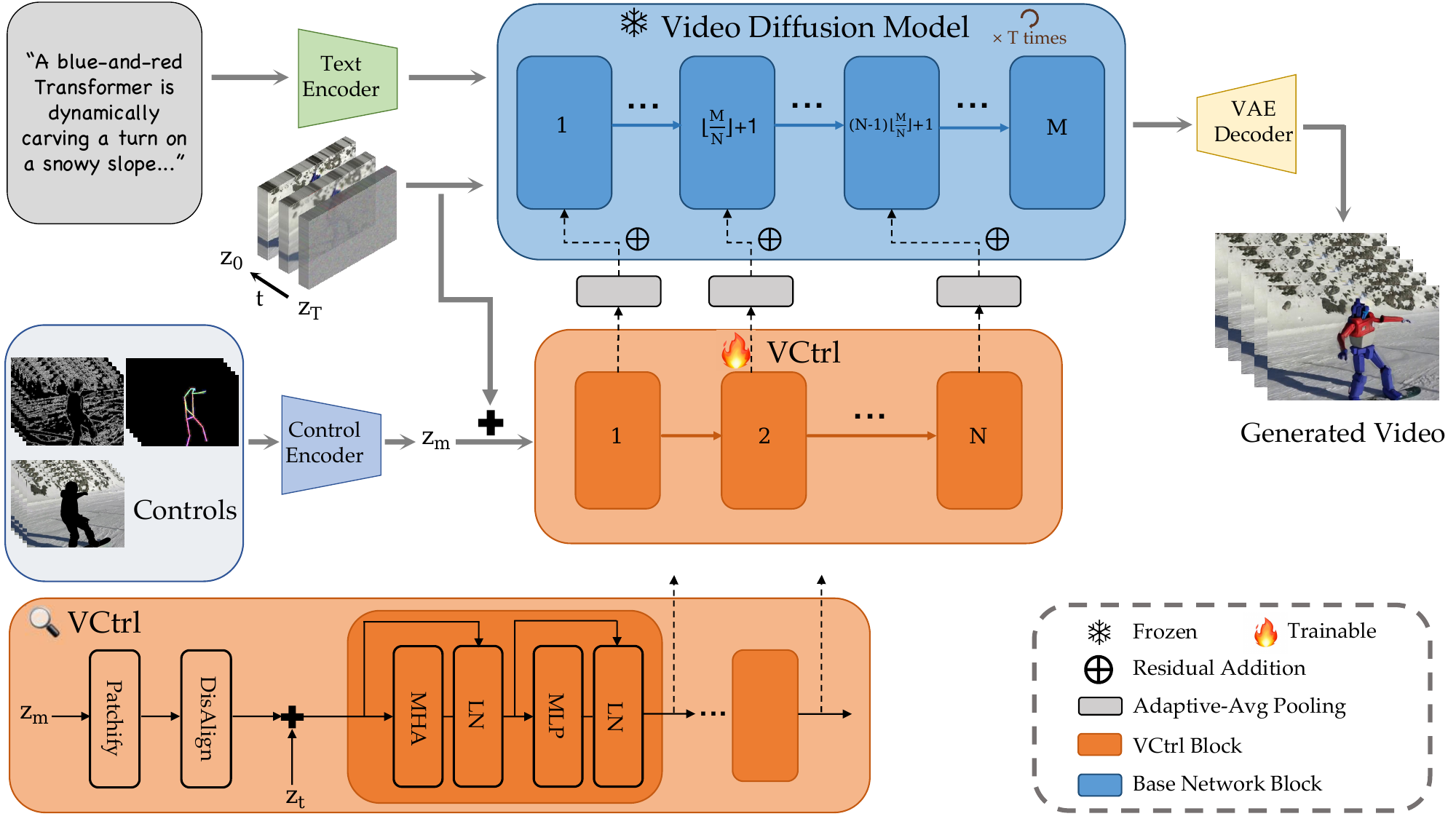}
    \vspace{-17pt}
    \caption{\textbf{Overview architecture of VCtrl.} A control signal (e.g., Canny edges, semantic masks, or pose keypoints) is first encoded by the control encoder. The resulting representation is then additively combined with latent and incorporated into the Video Diffusion Model via the proposed VCtrl module, which leverages a sparse residual connection mechanism. After several iterative denoising steps, the refined latent is decoded by a pretrained VAE to produce the final video.}
    \vspace{-7pt}
    \label{fig:overall}
\end{figure*}
\section{Related Work}
\label{sec:related}
\subsection{Video Diffusion Models}
Recent years have witnessed significant progress in text-to-video synthesis (T2V)~\cite{karras2023dreampose, ruan2023mm, zhang2023i2vgen, he2022latent, chen2023videocrafter1, hong2022cogvideo, girdhar2023emu, wang2023modelscope, an2023latent, gu2023reuse}. Research in this field can be broadly categorized into two main directions. The first direction~\cite{zhou2022magicvideo, blattmann2023align, guo2023animatediff, ge2023preserve, wang2023lavie} extends established text-to-image (T2I) frameworks by incorporating specialized components for temporal modeling. The second direction~\cite{ho2022video, ho2022imagen, singer2022make, yang2024cogvideox, kong2024hunyuanvideo} focuses on developing dedicated T2V frameworks trained from scratch. Despite these advancements, existing methods often rely on task-specific architectures and exhibit limited flexibility in handling diverse conditioning signals, which restricts their generalizability and practical applicability.

\subsection{Controllable Generation}
\label{subsec:controllable}

\para{Controllable Image Generation.} 
Advancements in image diffusion models have introduced sophisticated control mechanisms through architectural innovations and refined training strategies. The inherent properties of the diffusion process enable fundamental manipulation capabilities, such as color variation~\cite{meng2022sdedit} and region-specific inpainting~\cite{ramesh2022hierarchical}. For spatial control, ControlNet~\cite{zhang2023adding} proposes an innovative architecture that augments pre-trained models with spatial conditioning. Similarly, T2I-Adapters~\cite{mou2023t2i} achieve multi-condition control via lightweight feature alignment modules. Text-based control is realized through a combination of prompt engineering~\cite{brooks2022instructpix2pix}, CLIP feature manipulation~\cite{gal2022image}, and cross-attention modulation~\cite{hertz2022prompt}. These approaches collectively highlight the potential of unified architectures to handle diverse control signals while maintaining model efficiency

\para{Controllable Video Generation.} 
Several works~\cite{yang2024cogvideox,hong2022cogvideo,kong2024hunyuanvideo} have explored text-based guidance for conditional video generation. However, they often lack fine-grained controls. To address this limitation, recent research has shifted towards integrating additional conditions into video diffusion models. For instance, several studies~\cite{ma2023follow,xu2023magicanimate,tu2024stableanimator} focus on generating videos conditioned on sequences of human pose maps and reference images. Furthermore, DRAGNUWA~\cite{yin2023dragnuwa} and 3DTrajMaster~\cite{fu20243dtrajmaster} introduce trajectory information to enable fine-grained temporal control. Despite their impressive results, these models often rely on complex condition encoding schemes and domain-specific training strategies. To overcome these challenges, methods such as Text2Video-Zero~\cite{text2video-zero}, Control-A-Video~\cite{chen2023controlavideo}, and VideoComposer~\cite{wang2023videocomposer} adapt text-to-image models for controllable video generation. 
However, these approaches primarily repurpose image generation models rather than directly leveraging architectures specifically designed for video generation, resulting in suboptimal temporal coherence and visual consistency.
To address these limitations, we propose a unified framework that supports versatile control conditions, demonstrating scalability and adaptability to more complex scenarios. Our approach is compatible with both text-to-video and image-to-video models, enabling a wide range of video generation tasks.

%% file: sec/3_method.tex
\section{Methodology}
\label{sec:method}

We propose a unified framework for controllable video generation. Given an input control signal (e.g., Canny edges, semantic masks, or pose keypoints) paired with text prompt, our approach enables precise spatiotemporal control in video synthesis while maintaining high visual fidelity.  The overall architecture is illustrated in Figure~\ref{fig:overall}.
This section is structured as follows: Section~\ref{sec:preliminaries} reviews essential preliminaries on diffusion models; Section~\ref{sec:unified_control} introduces the unified control signal encoding process; Section~\ref{sec:control} details the architecture of our proposed VCtrl module; Section~\ref{sec:sparse_residual} presents the sparse residual connection mechanism; Section~\ref{sec:train} outlines the training methodology; and Section~\ref{sec:data} describes the data filtering pipeline.

\subsection{Preliminaries}
\label{sec:preliminaries}
Diffusion models (DMs)~\cite{ho2020denoising, sohl2015deep} are a class of generative models. Latent Diffusion Models (LDMs)~\cite{rombach2022high} extend this framework by operating in a learned latent space. For video generation, given an input video $x \in \mathbb{R}^{F \times H \times W \times 3}$, we first encode it into a compressed latent representation using a pre-trained encoder $\mathcal{E}$:
\begin{equation}
    z = \mathcal{E}(x) \in \mathbb{R}^{f \times h \times w \times ch},
\end{equation}
where $f < F$, $h < H$, $w < W$ typically, and $ch$ represents the channel dimension in the latent space.
The diffusion process then occurs in this latent space through $T$ noise-addition steps, producing noisy latents $z_1, z_2, ..., z_T$, where noise is injected into $z$ to obtain a noise-corrupted latent $z_t$ following the defined noise schedule~\cite{ho2020denoising}.
The reverse process learns to denoise these latents using a network $\epsilon_\theta$ trained via:
\begin{equation}
    \mathcal{L} = \mathbb{E}_{z,t,c,\epsilon} \left[ \| \epsilon - \epsilon_\theta(z_t, t, c) \|^2_2 \right],
\end{equation}
where $t \in [1,N] $, $\epsilon \sim \mathcal{N}(0,I)$, and $c$ represents a conditioning vector that provides additional context or constraints to guide the denoising process.

\subsection{Unified Control Signal Encoding}
\label{sec:unified_control}
Existing models often incorporate control mechanisms tailored to specific tasks, limiting their generalizability. 
To address this, we propose a unified control signal encoding process capable of handling diverse control types.
Our method integrates control by utilizing control videos as the primary input for encoding control signals, enabling flexible adaptation to diverse controllable generation tasks.
This approach enables natural generalization across a wide range of control tasks. Subsequently, these signals are represented through a cohesive control signal encoding framework, referred to as the Control Encoder. 

The Control Encoder in VCtrl is designed to process these various control signals in a unified manner. It accepts a video-based control signal $v_c \in \mathbb{R}^{F \times H \times W \times 3}$ as input, where each frame represents a specific control signal at a given time step. This format naturally accommodates a wide range of control types. The control video is first encoded into a latent representation $z_c$ using a pre-trained variational autoencoder $\mathcal{E}$:
\begin{equation}
    z_c = \mathcal{E}(v_c) \in \mathbb{R}^{f \times h \times w \times ch}.
\end{equation}
To further enhance adaptability, we incorporate a task-aware mask sequence $M_c \in \{0, 1\}^{f \times h \times w}$ alongside the input conditions. 
For Canny edge and human pose control, this mask indicates whether each frame is conditioned, while for segmentation mask control, it indicates the segmented area. 
We then concatenate this mask along the channel dimension with the encoded information $z_c$, resulting in a combined representation that captures both the latent features and the task-aware control signals. This representation, denoted as:
\begin{equation}
    z_{m} = z_c \oplus M_c,
    \label{eq:z_control}
    \end{equation}
thereby enabling a more effective representation of various types of control inputs and enhancing the model's adaptability across different tasks.

\subsection{VCtrl Module} 
\label{sec:control}
In line with~\cite{zhang2023adding}, we define the term {\em network block} as a collection of neural layers that are typically combined to create a single unit within a neural network, e.g., resnet block, conv-bn-relu block, multi-head attention block, transformer block, etc. For an input feature map $\bm{x}_i$, the output feature map of the $i$-th block $\mathcal{F}^i$ is calculated as follows, where $\Theta^i$ denotes the parameters associated with that block:
\begin{equation}
    \bm{y}^i = \mathcal{F}^i(\bm{x}^i; \Theta^i) 
    \label{eq:output_feature_map}.
\end{equation}

In the context of VCtrl, we define a base network consisting of $M$ total blocks, represented as $\mathcal{F}_b^i(\cdot;\Theta_{\text{b}}^i)$ for each block.
We freeze the parameters of all blocks in the base model, while introducing parallel  sub-network, referred to as the VCtrl module, with trainable parameters $\Theta_c^i$ at $N$ control points selected at fixed intervals, represented as $\mathcal{F}_c^i(\cdot;\Theta_{\text{c}}^i)$ for each block.
In this study, we utilize CogVideoX~\cite{hong2022cogvideo} as an example to illustrate the capability of VCtrl in augmenting conditional control within a large pretrained video diffusion model.

The VCtrl module is predicated on a lightweight Transformer Encoder architecture, as depicted in orange in Figure~\ref{fig:overall}. Its primary aim is to proficiently receive and process a variety of input modalities through a series of compact blocks. 
This architecture seamlessly integrates the initial feature map \(\bm{x}_0\) from the base network with control information extracted from an externally sampled conditioning signal \(\bm{z_m}\), thereby encoding temporal features and enhancing the model's capacity to capture intricate temporal relationships through a multi-input framework.
By design, the VCtrl comprises approximately one-fifth the number of blocks relative to the base network, represented as \(\mathcal{F}_c^i(\cdot;\Theta_{\text{c}}^i)\) for each block. 
To ensure precise alignment of control signals with latent representations, we propose a DistAlign layer, which adaptively scales the control signals to match latent dimensions. This approach effectively mitigates noise interference arising from discrepancies in signal scales, enhancing the stability and consistency of the generative process.

\subsection{Sparse Residual Connection Mechanism}
\label{sec:sparse_residual}
To integrate external conditioning information while preserving the stability of large pre-trained models, we propose a sparse residual connection mechanism that injects control signals via parallel, lightweight VCtrl sub-networks. 
In our approach, the base network's parameters remain completely frozen, and the control information is introduced at fixed intervals through additional trainable branches.

Let the base network consist of $M$ blocks, and suppose we select $N$ control points. We define the indices of the blocks of the base network where the control branches are attached as follows:

\begin{equation}
    \mathcal{I} = \left\{ (k - 1) \cdot \left\lfloor \frac{M}{N} \right\rfloor + 1 \right\}_{k=1}^{N}.
\end{equation}

At each control point $i \in \mathcal{I}$, the output of the base network block is given by $y_{b}^i = \mathcal{F}_b^i(x_{b}^i;\Theta_{\text{b}}^i)$, as defined in Equation~\ref{eq:output_feature_map}.
In parallel, a VCtrl sub-network processes the same input along with an external conditioning vector $\bm{z_m}$, yielding $y_{c}^i = \mathcal{F}_c^i(x_{c}^i;\Theta_{\text{c}}^i)$.

To reconcile potential differences in the spatial and temporal dimensions between $y_{b}^i$ and $y_{c}^i$, an adaptive average pooling operation, denoted by $\mathrm{AdaptiveAvgPool}(\cdot)$, is applied to $y_{c}^i$. 
Specifically, adaptive average pooling automatically adjusts the hidden dimensions of the feature map $y_{c}^i$ to exactly match those of $y_{b}^i$, ensuring compatibility for feature fusion. 
The final output at control point $i$, represents the input feature map of the next block in the base network, is then obtained through a residual fusion:
\begin{equation}
    x_{b}^{i+1} = y_{b}^i + \mathrm{AdaptiveAvgPool}(y_{c}^i).
\end{equation}

The proposed mechanism employs sparse alignment and residual fusion to enhance generative capabilities. Sparse alignment maintains a one-to-one correspondence between VCtrl blocks and base network layers, ensuring balanced injection of control signals while preserving hierarchical structure. Residual fusion utilizes adaptive average pooling to merge control signals effectively with the original features. By freezing the base network and training lightweight VCtrl sub-networks, our method integrates control signals efficiently across Transformer layers with minimal computational overhead.

\subsection{Training}
\label{sec:train}
Given an input video $x \in \mathbb{R}^{F \times H \times W \times 3}$, we first encode it into a compact latent representation $z$ using a pretrained encoder. We progressively introduce noise into this latent representation through $T$ iterative steps, obtaining the noisy latent $z_t$ at each timestep $t$ according to a predefined noise schedule. To achieve external controllability in video generation, we introduce new video conditioning signals $z_m$(see Equation~\ref{eq:z_control}), while also utilizing existing signals $c$, such as textual prompt or reference frame, derived from the base network.

During training, we freeze the parameters of the pretrained base diffusion model and solely optimize the VCtrl module. The controllable denoising network $\epsilon_\theta$ learns to predict the noise added at each timestep guided by these conditioning inputs. We consider $\epsilon$ to be drawn from a normal distribution $\epsilon \sim \mathcal{N}(0, I)$:
\begin{equation}
    \mathcal{L}_{\text{VCtrl}} = \mathbb{E}_{z, t, c, z_m, \epsilon} \left[\|\epsilon - \epsilon_\theta(z_t, t, c, z_m)\|_2^2\right],
\end{equation}
where $\epsilon \sim \mathcal{N}(0,I)$, $\mathcal{L}$ represents the training objective for the controllable video diffusion model. This loss directly guides the finetuning process of the diffusion model with the proposed VCtrl module, enabling effective generation of videos that adhere to specified textual and control video constraints.

\begin{figure}
    \centering
    \includegraphics[width=0.8\linewidth]{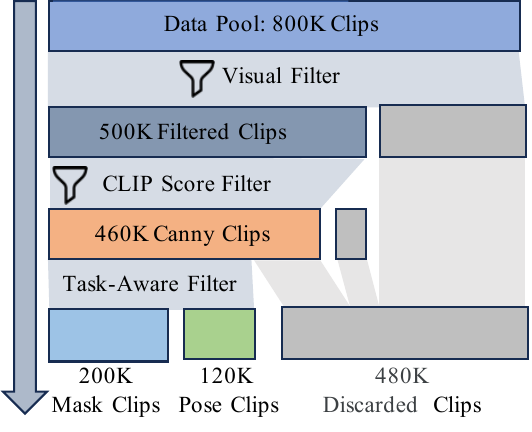}
    \vspace{-7pt} 
    \caption{\textbf{Our Data Filtering Pipeline.} Videos refined by an aesthetic filter are recaptioned and processed to extract Canny edges, human keypoints, and segmentation masks, providing training data for diverse controllable tasks.}
    \label{fig:datafilter}
    \vspace{-17pt}
\end{figure}

\subsection{Data}
\label{sec:data}
We utilize three publicly available video datasets—WebVid-10M, MiraData~\cite{ju2025miradata}, and Vript~\cite{yang2025vript}—to form an initial corpus comprising approximately 800K text-video pairs. After systematic data processing and filtering, we generate three datasets specific to canny, mask, and human posture control tasks respectively.

To ensure high-quality training data, we apply a hierarchical filtering pipeline illustrated in Figure~\ref{fig:datafilter}. Specifically, we conduct the following filtering steps: 1) Visual Filter: We perform visual filtering, including scene segmentation, border removal, and aesthetic filtering~\cite{wu2023towards}. Scene segmentation divides the original video into segments by comparing hash values of consecutive frames. Black borders are identified and removed based on the standard deviation of the color histogram. Subsequently, aesthetic filters exclude low-quality frames, resulting in video clips with resolutions ranging from $446 \times 336$ to $1280 \times 720$, each containing a maximum of 160 frames. 2) CLIP Score Filter: We employ a recaptioning model~\cite{hong2024cogvlm2} to regenerate detailed and accurate captions for aesthetically filtered videos. To ensure semantic relevance, we calculate CLIP scores comparing both original and regenerated captions against their respective videos, retaining captions above a defined quality threshold. 3) Task-Aware Filter: The refined videos undergo task-aware preprocessing for conditional video generation, including: \textbf{Canny edge detection} using hysteresis thresholds and Gaussian smoothing ($\sigma=1.0$)\cite{canny1986computational}; \textbf{semantic mask extraction} performed via segmentation model\cite{ravi2024sam}, maintaining consistent video-level segmentation and incorporating dynamic multi-target masking and random dilation for robustness; and \textbf{human pose estimation} using a pose extraction model~\cite{xu2022vitpose} that detects 133 keypoints per individual visualized against a uniform background, with temporal smoothing applied to ensure motion consistency.

%% file: sec/4_experiments.tex
\section{Experiments}
\label{sec:exp}

\begin{figure*}
    \vspace{-3pt}
    \centering
    \includegraphics[width=1\linewidth]{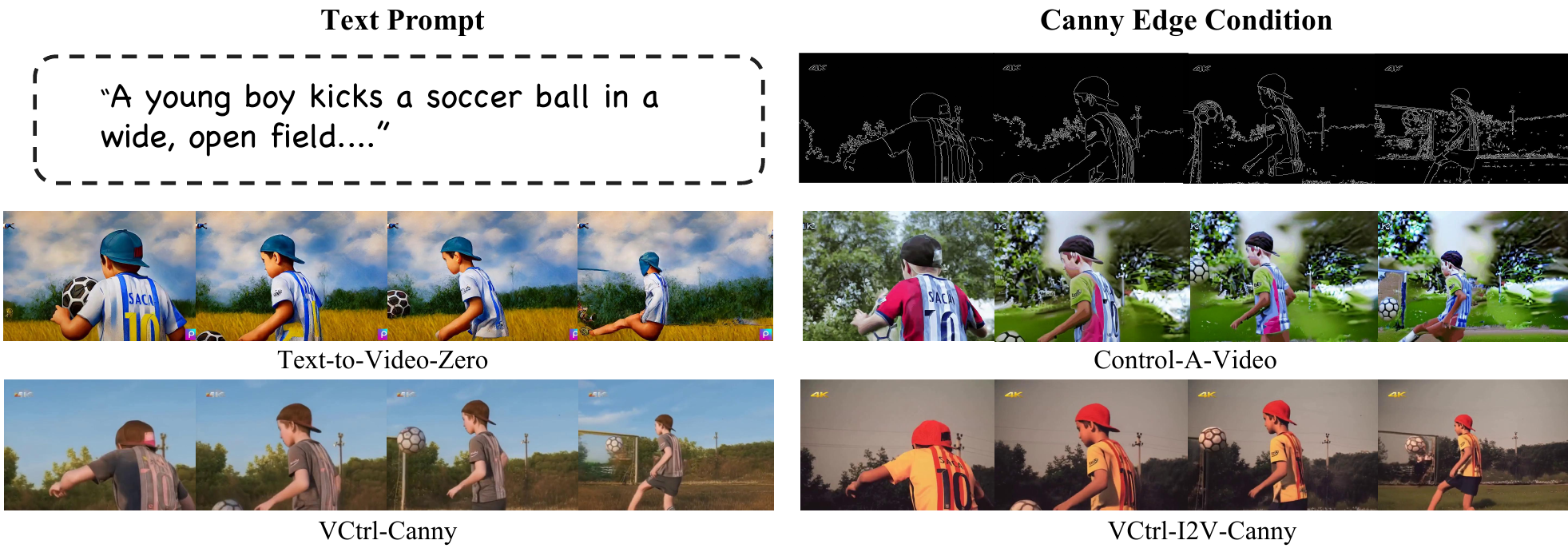}
    \vspace{-20pt}
    \caption{\textbf{Qualitative comparison to previous methods.} We compare our method with Control-A-Video~\cite{chen2023controlavideo} and Text2Video-Zero~\cite{text2video-zero}, demonstrating superior visual coherence and stronger adherence to the Canny edge conditions.}
    \vspace{-7pt}
    \label{fig:canny_comparision}
\end{figure*}

\subsection{Implementation Details}
We implement VCtrl using a generalizable video diffusion architecture compatible with various block-structured base networks for video generation. Due to its superior temporal coherence and strong scalability, in this work, we specifically select CogVideoX-5B~\cite{yang2024cogvideox} as our primary base network, along with its I2V variant (CogVideoX-5B-I2V), which additionally accepts an extra reference frame as input.
All input videos, paired with their corresponding condition videos, are set to a resolution of $720\times480$ or $480\times720$, and 49 consecutive frames are extracted from each video pair to be used as training data.
We set the learning rate to $1\times10^{-5}$ and employ the Adam optimizer with $\beta_1=0.9$, $\beta_2=0.999$, applying gradient clipping with a maximum norm of 1.0 to ensure training stability.
To further enhance robustness, we employ a truncated normal distribution-based random cropping method, which adaptively selects the cropping center and boundaries based on the video's aspect ratio. The standard deviation is set to 0.25 of the maximum allowable offset in height or width.

\subsection{Evaluation Metrics}
Our evaluation framework addresses two primary dimensions: video quality assessment and control precision analysis. For video quality, we adopt Fréchet Video Distance (FVD)~\cite{unterthiner2018towards}, Subject Consistency\cite{liu2023evalcrafter}, and Aesthetic Score~\cite{ke2021musiq}.
Due to the absence of standardized metrics for evaluating control precision, we propose three novel metrics inspired by prior works in controllable image and video generation~\cite{li2021image, ju2023humansd, wang2024motionctrl}, specifically tailored for video generation conditioned on Canny edge maps (\textbf{Canny-to-Video}), binary subject masks (\textbf{Mask-to-Video}), and human keypoint sequences (\textbf{Pose-to-Video}):

\para{Canny Matching.}
Given the ground-truth edge sequence $\{C^{gt}_i\}_{i=1}^F$ and the generated edge sequence $\{C^{pred}_i\}_{i=1}^F$ , both extracted using Canny edge detector~\cite{canny1986computational}, respectively. 
We then binarize the obtained edge maps into binary values. 
Subsequently, edge alignment is quantified using an adaptive Dice coefficient~\cite{sudre2017generalised} defined as follows:
\begin{equation}
    \mathcal{S}_{\text{canny}} = \frac{2}{F}\sum_{i=1}^{F} \frac{|C^{pred}_i \cap C^{gt}_i| + \epsilon}{|C^{pred}_i| + |C^{gt}_i| + \epsilon}
\end{equation}
where $\epsilon=1e^{-5}$ prevents division by zero, and $|\cdot|$ denotes pixel count.

\para{Masked Subject Consistency (MS-Consistency).}
Given binary subject masks $\{M_i\}_{i=1}^F$, we quantify the consistency between the generated video $V^{pred}$ and ground truth $V^{gt}$ videos by calculating the RGB pixel-wise L1 distance:
\begin{equation}
    S_{mask} =\sum^F_{i=1}\frac{|| M_{i}(V^{pred}_{i} - V^{gt}_{i})||_1}{||M_i||_1}
\end{equation}
where $|| \cdot ||_1$ indicates the L1 norm.

\para{Pose Similarity.}
Adopting VitPose~\cite{xu2022vitpose} detector, we compute Object Keypoint Similarity (OKS) ~\cite{xiao2018simple} between generated pose sequences ${p}_k^{pred}$ and ground truth pose sequences ${p}_k^{gt}$:
\begin{equation}
    \mathcal{S}_{\text{pose}} = \frac{1}{F}\sum_{i=1}^{F} \frac{1}{K}\sum_{k=1}^{K} \exp\left(-\frac{\|\mathbf{p}_k^{pred}-\mathbf{p}_k^{gt}\|^2_2}{2\sigma_k^2 A_i}\right)
\end{equation}
where $\sigma_k$ denotes keypoint-specific tolerance, $A_i$ is bounding box area, and $K$ represents the total number of keypoints (17).

\subsection{Qualitative Evaluation}
For a comprehensive qualitative evaluation, we examine our approach from two primary perspectives. First, we illustrate that the proposed VCtrl-I2V model can address diverse video generation tasks. As depicted in Figure \ref{fig:summary}, our method successfully supports style transfer (Canny), video editing (Mask), and character animation (Pose), consistently producing high-fidelity video content despite substantial motion. Crucially, it preserves spatiotemporal coherence across frames, ensuring smooth and temporally consistent transitions. 
Second, in Figure~\ref{fig:canny_comparision}, we visually compare our models against representative baselines.
Control-A-Video~\cite{chen2023controlavideo} generates videos with temporal inconsistencies, exhibiting abrupt content changes between frames. Text2Video-Zero~\cite{text2video-zero} maintains better control adherence but suffers from background artifacts and low quality subject rendering. 
Our VCtrl-Canny achieves strict canny constraint satisfaction while preserving visual fidelity. The I2V-enhanced version further improves temporal coherence (evident in stable color profiles) and maintains first-frame fidelity throughout the sequence.

\begin{table}[!t]
    \centering
    \resizebox{0.9\linewidth}{!}{
    \begin{tabular}{l c c}
    \toprule
    Model & Canny Matching $\uparrow$ & FVD $\downarrow$ \\
    \midrule
    CogVideoX-5B~\cite{yang2024cogvideox} & - & 1596.51 \\
    CogVideoX-5B-I2V~\cite{yang2024cogvideox} & - & 989.32 \\
    \midrule
    Text2Video-Zero~\cite{text2video-zero} & 0.20 & 1761.82 \\
    Control-A-Video~\cite{chen2023controlavideo} & 0.14 & 1298.26 \\
    \midrule
    VCtrl-Canny & 0.24 & 985.31 \\
    VCtrl-I2V-Canny & \textbf{0.28} & \textbf{345.00} \\
    \bottomrule
    \end{tabular}}
    \vspace{-7pt}
    \caption{\textbf{Quantitative evaluation for Canny-to-Video generation.} We report Canny Matching for control effectiveness (higher is better) and FVD for video quality (lower is better).}
    \vspace{-7pt}
    \label{tab:canny_eval}
\end{table}

\begin{table}[!t]
    \centering
    \resizebox{0.85\linewidth}{!}{
    \begin{tabular}{l c c}
    \toprule
    Model & MS-Consist. $\uparrow$ & FVD $\downarrow$ \\
    \midrule
    CogVideoX-5B~\cite{yang2024cogvideox} & - & 1592.88 \\
    CogVideoX-5B-I2V~\cite{yang2024cogvideox} & - & 1132.28 \\
    \midrule
    CoCoCo~\cite{zi2024cococo} & 0.32 & 961.17 \\
    \midrule
    VCtrl-Mask & 0.36 & 480.86 \\
    VCtrl-I2V-Mask & \textbf{0.63} & \textbf{228.78} \\
    \bottomrule
    \end{tabular}}
    \vspace{-7pt}
    \caption{\textbf{Quantitative evaluation for Mask-to-Video generation.} We report MS-Consistency (higher is better) for control effectiveness and FVD (lower is better) for video quality.}
    \vspace{-7pt}
    \label{tab:mask_eval}
\end{table}

\subsection{Quantitative Evaluation}
We present a comprehensive quantitative evaluation of our methods against existing representative approaches across three video generation tasks. For each task, we select suitable benchmarks and both established and newly proposed metrics to ensure a thorough comparison.

\para{Canny-to-Video.} 
We quantitatively evaluate our approach on the Canny-to-Video generation task using videos derived from the Davis dataset~\cite{Perazzi_CVPR_2016}, where ground-truth edges are extracted using Canny edge detection.
The quantitative results presented in Table~\ref{tab:canny_eval} demonstrate that our VCtrl-based methods surpass existing methods in both control precision and visual quality. 
Specifically, VCtrl-Canny and VCtrl-I2V-Canny achieve improvements of 0.04 and 0.08, respectively, over Text2Video-Zero~\cite{text2video-zero} in terms of the Canny Matching metric.
Regarding visual quality measured by the FVD score, VCtrl-Canny and VCtrl-I2V-Canny reduce scores by roughly 292 and 313 points, respectively, compared to Text2Video-Zero, and by 611 and 644 points compared to their corresponding base models (CogVideoX-5B and CogVideoX-5B-I2V, respectively).

\begin{table}[t]
    \centering
    \resizebox{0.9\linewidth}{!}{
    \begin{tabular}{l c c c }
    \toprule
    Model  & Pose Similarity $\uparrow$ & FVD $\downarrow$ \\
    \midrule
    CogVideoX-5B-I2V~\cite{yang2024cogvideox} & 0.60 & 837.44 \\
    \midrule
    Moore-AnimateAnyone~\cite{moorethreads2024} & 0.82 & 702.59 \\
    ControlNeXt-SVD~\cite{peng2024controlnext} & 0.82 & 255.50 \\
    \midrule
    VCtrl-I2V-Pose & \textbf{0.98} & \textbf{175.20} \\
    \bottomrule
    \end{tabular}}
    \vspace{-7pt}
    \caption{\textbf{Quantitative evaluation for Pose-to-Video generation.} We report  pose matching for control effectiveness and FVD for video quality.}
    \vspace{-7pt}
    \label{tab:pose_eval}
\end{table}
\para{Mask-to-Video.} 
We quantitatively evaluate our approach on Mask-to-Video generation using a dataset derived from Davis~\cite{Perazzi_CVPR_2016}, where ground-truth masks are extracted using semantic mask extraction. As summarized in Table~\ref{tab:mask_eval}, our proposed methods consistently outperform existing approaches. 
Specifically, VCtrl-Mask and VCtrl-I2V-Mask achieve improvements of 0.04 and 0.31 in Masked Subject Consistency and reduce the FVD scores by approximately 480 and 732 points, respectively, compared to CoCoCo~\cite{zi2024cococo}.

\begin{table*}[!t]
    \centering
    \resizebox{0.85\linewidth}{!}{
    \begin{tabular}{l l c c c c c c}
    \toprule
    \multirow{2}{*}{Task} & \multirow{2}{*}{Model} & Overall & Temporal & Text & Facial Identity & Pose & Background \\
    & & Quality & Consist. & Alignment & Consist. & Consist. & Consist. \\
    \midrule
    \multirow{4}{*}{Canny-to-Video} & Control-A-Video~\cite{chen2023controlavideo} & 1.47 & 1.52 & 2.26 & - & - & - \\
                           & Text2Video-Zero~\cite{text2video-zero} & 1.44 & 1.27 & 2.38 & - & - & - \\
                           \cmidrule{2-8}
                           & VCtrl-Canny & \textbf{2.96} & \textbf{3.13} & \textbf{3.42} & - & - & - \\
    \midrule
    \multirow{2}{*}{Mask-to-Video} & CoCoCo~\cite{zi2024cococo} & 2.05 & 1.90 & 2.21 & - & - & 2.50 \\
                           \cmidrule{2-8}
                           & VCtrl-Mask & \textbf{3.04} & \textbf{3.33} & \textbf{3.18} & - & - & \textbf{3.26} \\
    \midrule
    \multirow{3}{*}{Pose-to-Video}  & Moore-AnimateAnyone~\cite{moorethreads2024} & 1.38 & 1.25 & - & 1.26 & 1.36 & - \\
                           & ControlNeXt~\cite{peng2024controlnext} & 2.85 & 2.71 & - & 2.50 & 3.04 & - \\
                           \cmidrule{2-8}
                           & VCtrl-I2V-Pose & \textbf{3.30} & \textbf{3.21} & - & \textbf{3.06} & \textbf{3.39} & - \\
    \bottomrule
    \end{tabular}}
    \vspace{-7pt}
    \caption{\textbf{User study comparing VCtrl with competing methods.} All methods are evaluated using identical inputs for each task, with scores ranging from 1 (lowest) to 5 (highest).}
    \vspace{-7pt}
    \label{tab:user_study}
\end{table*}

\begin{table}
    \centering
    \resizebox{\linewidth}{!}{
    \begin{tabular}{l c c c c}
    \toprule
    Layout
        & FVD~$\downarrow$
        & Subject Consist.~$\uparrow$
        & Aesthetic Score~$\uparrow$
        & Canny Matching~$\uparrow$ \\
    \midrule
    Even & 1005.98 & 0.880 & 0.450 & 0.226  \\
    End & 1449.24 & 0.847 & 0.450 & 0.124 \\
    Space & \textbf{949.46} & \textbf{0.884} & \textbf{0.473} &\textbf{0.248} \\
    \bottomrule
    \end{tabular}}
    \vspace{-7pt}
    \caption{\textbf{Comparison of control layout designs across multiple metrics.} The Space layout achieves superior overall performance, demonstrated by higher visual quality scores, improved Canny matching, and a lower FVD score.}
    \vspace{-12pt}
    \label{tab:model_size_comparison}
\end{table}

\begin{table}
    \centering
    \resizebox{\linewidth}{!}{
    \begin{tabular}{l c c c c}
    \toprule
    Model 
        & FVD~$\downarrow$
        & Subject Consist.~$\uparrow$
        & Aesthetic Score~$\uparrow$
        & Canny Matching~$\uparrow$ \\
    \midrule
    VCtrl-Small  & 1001.75 & 0.882 & 0.459 & 0.205 \\
    VCtrl-Medium & 949.46 & 0.884 & \textbf{0.473} & \textbf{0.248} \\
    VCtrl-Large & \textbf{937.37} & \textbf{0.889} & 0.471 & 0.231 \\
    \bottomrule
    \end{tabular}}
    \vspace{-7pt}
    \caption{\textbf{Comparison of VCtrl variants with different model complexities.} Models of varying sizes are evaluated comprehensively on visual quality, subject consistency, aesthetic score, and control precision.}
    \vspace{-15pt}
    \label{tab:layout_comparison}
\end{table}

\para{Pose-to-Video.} 
We quantitatively evaluate our approach on the Pose-to-Video generation task using an evaluation set of 100 videos selected from publicly available datasets~\cite{zablotskaia2019dwnet, siarohin2019first, siarohin2021motion, jafarian2021learning}, where ground-truth poses are extracted using human pose estimation. As summarized in Table~\ref{tab:pose_eval}, our proposed method consistently outperforms existing approaches. Specifically, VCtrl-I2V-Pose achieves a significant improvement of approximately 0.16 over  in Pose Similarity and  reductions of roughly 80 points in the FVD score compared to ControlNeXt-SVD\,\cite{peng2024controlnext}.

Despite employing a relatively simple architecture without sophisticated modules utilized in prior domain-specific methods~\cite{ma2023follow,xu2023magicanimate,zi2024cococo,yin2023dragnuwa}, our comprehensive evaluations demonstrate that VCtrl consistently achieves competitive or superior performance, significantly enhancing the generation capabilities of the base models. 
Moreover, improvements observed across established video quality metrics such as FVD are consistently corroborated by our newly proposed metrics (Canny Matching, MS-Consistency, and Pose Similarity), which provide a more intuitive and precise measure of control effectiveness. 
These results underscore the effectiveness, efficiency, and adaptability of VCtrl in diverse controllable video generation scenarios.

\subsection{Ablative Study}
\begin{figure}[t]
    \includegraphics[width=\linewidth]{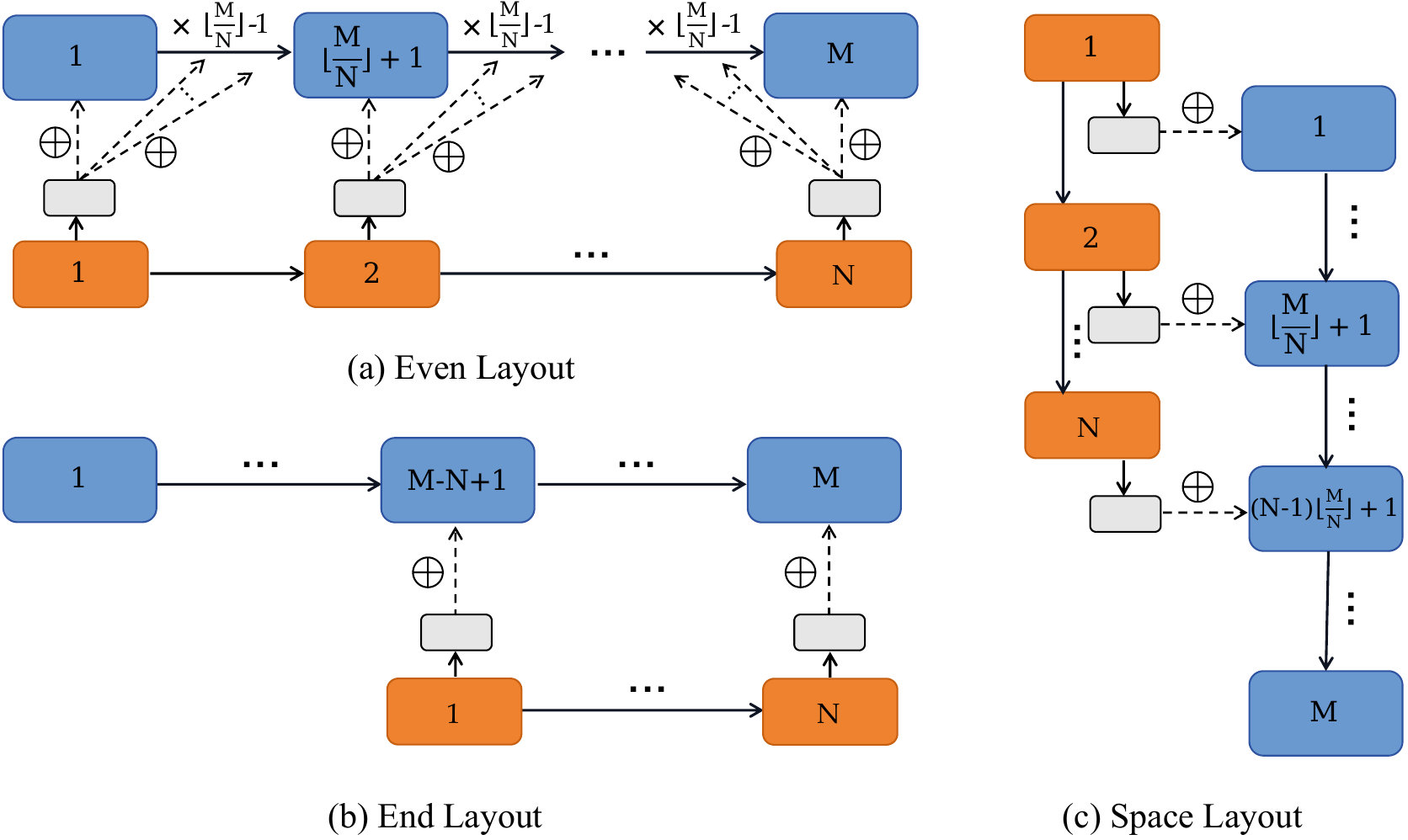}
    \vspace{-17pt}
    \caption{\textbf{Control Layouts.} (a) Even: control signals uniformly injected throughout the network; (b) End: control signals densely injected toward the end of the network; (c) Space: control signals sparsely and evenly distributed across the network.}
    \vspace{-7pt}
    \label{fig:layout}
\end{figure}

\para{Connection Layout Design.}
We investigate architectural variations of VCtrl by evaluating three distinct control block connection layouts: even, end, and space. These layouts explore different strategies for integrating control signals using the sparse residual connection mechanism introduced in Section \ref{sec:sparse_residual}. Figure~\ref{fig:layout} illustrates the conceptual differences among these layouts. 
To ensure a fair evaluation, each variant is trained under identical conditions for 35,000 optimization steps on the Canny-to-Video task. 
Comparing results in Table~\ref{tab:model_size_comparison}, the space layout consistently delivers better performance, despite employing sparser integration of control signals compared to the even layout. Conversely, concentrating control blocks exclusively toward the end of the network yields the weakest results, suggesting that distributed integration of control signals is essential for optimal performance.

\para{Complexity.}
To systematically investigate the balance between computational complexity and control performance, we conduct experiments with three variants of VCtrl, varying the ratio of VCtrl blocks to base network blocks: VCtrl-Small (1:15), VCtrl-Medium (1:5), and VCtrl-Large (1:2). 
Each variant was trained for an identical total of 35,000 optimization steps to ensure a fair comparison.
Table~\ref{tab:layout_comparison} presents a quantitative comparison across multiple metrics in video generation task guided by Canny edges. 
Despite significant parameter reductions compared to the base network and VCtrl-Large, VCtrl-Medium maintains robust performance.
This suggests that lightweight models can maintain control effectiveness while enhancing computational efficiency, which is crucial for many real-world applications. 
Thus, we adopt VCtrl-Medium as our primary model variant for the Canny, Pose, and Mask tasks, as it achieves an optimal trade-off between model complexity and performance.

\subsection{User Study} 
We conducted a user study to quantitatively evaluate our proposed methods against established baselines on three conditional video generation tasks. For each task, we selected 20 representative video samples, which were independently assessed by domain experts in a blind evaluation setting. Participants rated each video on a scale from 1 (lowest) to 5 (highest) across multiple task-aware criteria, including Overall Quality, Temporal Consistency, Text Alignment, Facial Identity Consistency, Pose Consistency, and Background Consistency. The detailed results are summarized in Table~\ref{tab:user_study}. Our proposed methods consistently outperform existing baselines: VCtrl-Canny shows superior overall quality and temporal consistency; VCtrl-Mask significantly surpasses CoCoCo~\cite{zi2024cococo} in overall and background consistency metrics; and VCtrl-I2V-Pose notably improves overall quality, temporal coherence, facial identity, and pose consistency. These findings underscore the effectiveness and robustness of our proposed approaches for versertile controllable video generation scenarios.

%% file: sec/5_conclusion.tex
\section{Conclusion}
\label{sec:conclusion}

We introduce a unified framework for controllable video generation, effectively integrating diverse controls through a unified control signal encoding strategy and a generalizable conditional module. Our sparse residual connection mechanism seamlessly incorporates these unified representations into pretrained video diffusion models, enabling precise and flexible video synthesis. Extensive experiments validate our framework’s effectiveness across various controllable generation tasks, demonstrated through quantitative evaluations and human assessments. Additionally, the lightweight and modular design ensures broad compatibility, facilitating future adaptation to a wider range of video generation architectures and applications.